\documentclass[sigconf]{acmart}

\usepackage{booktabs} % For formal tables

\setcopyright{rightsretained}
\copyrightyear{2017}
\acmYear{2017}
\setcopyright{acmcopyright}
\acmConference{CIKM'17 }{November 6--10, 2017}{Singapore, Singapore}\acmPrice{15.00}\acmDOI{10.1145/3132847.3133127}
\acmISBN{978-1-4503-4918-5/17/11}

\begin{document}
\title{Hybrid MemNet for Extractive Summarization}
%\titlenote{Produces the permission block, and
%  copyright information}
%\subtitle{Extended Abstract}
%\subtitlenote{The full version of the author's guide is available as
%  \texttt{acmart.pdf} document}

\author{Abhishek Kumar Singh}
%\authornote{Dr.~Trovato insisted his name be first.}
\affiliation{%
  \institution{IIIT Hyderabad}
  \streetaddress{}
  \city{} 
  \state{} 
  \postcode{}
}
\email{abhishek.singh@research.iiit.ac.in}

\author{Manish Gupta}
%\authornote{Dr.~Trovato insisted his name be first.}
\affiliation{%
  \institution{IIIT Hyderabad}
  \streetaddress{}
  \city{} 
  \state{} 
  \postcode{}
}
\authornote{The author is also a Principal Applied Researcher at Microsoft.}
\email{manish.gupta@iiit.ac.in}

\author{Vasudeva Varma}
%\authornote{Dr.~Trovato insisted his name be first.}
\affiliation{%
  \institution{IIIT Hyderabad}
  \streetaddress{}
  \city{} 
  \state{} 
  \postcode{}
}
\email{vv@iiit.ac.in}

\renewcommand{\shortauthors}{}

\begin{abstract}
Extractive text summarization has been an extensive research problem in the field of natural language understanding. While the conventional approaches rely mostly on manually compiled features to generate the summary, few attempts have been made in developing data-driven systems for extractive summarization. To this end, we present a fully data-driven end-to-end deep network which we call as Hybrid MemNet for single document summarization task. The network learns the continuous unified representation of a document before generating its summary. It jointly captures local and global sentential information along with the notion of summary worthy sentences. Experimental results on two different corpora confirm that our model shows significant performance gains compared with the state-of-the-art baselines.
\end{abstract}


%\ccsdesc[100]{Information systems}	


\maketitle

\section{Introduction}

The tremendous growth of the data over the web has increased the need to retrieve, analyze and understand a large amount of information, which often can be time-consuming. Motivation to make a concise representation of large text while retaining the core meaning of the original text has led to the development of various summarization systems. Summarization methods can be broadly classified into two categories: \textit{extractive} and \textit{abstractive}. Extractive methods aim to select salient phrases, sentences or elements from the text while abstractive techniques focus on generating summaries from scratch without the constraint of reusing phrases from the original text.

Most successful summarization systems use extractive methods. Sentence extraction is a crucial step in such systems. The idea is to find a representative subset of sentences, which contains the information of the entire set. Traditional approaches to extractive summarization identify sentences based on human-crafted features such as sentence position and length~\cite{erkan2004lexrank}, the words in the title, the presence of proper nouns, content features like term frequency~\cite{nenkova2006compositional}, and event features like action nouns~\cite{filatova2004event}. Generally, sentences are assigned a saliency score indicating the strength of presence of these features. Kupiec et al.~\shortcite{kupiec1995trainable} use binary classifiers to select summary worthy sentences. Conroy and O'Leary~\shortcite{conroy2001text} investigated the use of Hidden Markov Models while ~\cite{erkan2004lexrank,mihalcea2005language} introduced graph-based algorithms for selecting salient sentences.

Recently, interest has shifted towards neural network based approaches for modeling the extractive summarization task. Kageback et al.~\shortcite{kaageback2014extractive} employed the recursive autoencoder~\cite{socher2011dynamic} to summarize documents. Yin and  Pei~\shortcite{yin2015optimizing} exploit convolutional neural networks to project sentences to a continuous vector space and select sentences based on their `prestige' and `diversity' cost for the multi-document extractive summarization task. Very recently, Cheng and Lapata~\shortcite{cheng2016neural} introduced attention based neural encoder-decoder model for extractive single document summarization task, trained on a large corpus of news articles collected from Daily Mail. Similar to Cheng and Lapta~\shortcite{cheng2016neural}, our work is focused on sentential extractive summaries of single document using deep neural networks. However, we propose the use of memory networks and convolutional bidirectional long short term memory networks for capturing better document representation.

\begin{figure}[t]
%\small
%[height=2.5in, width=5in]
\includegraphics[height=2.7in]{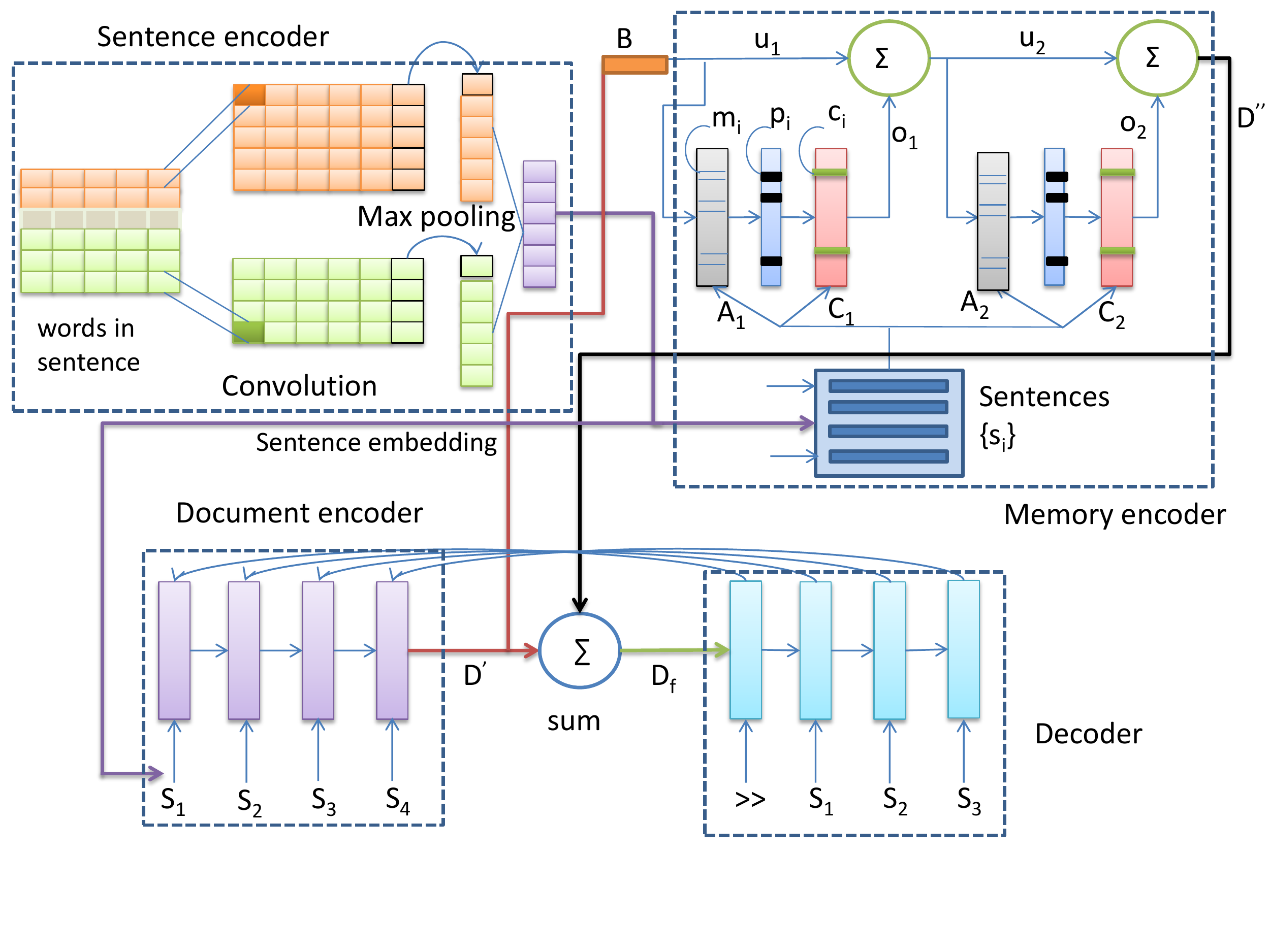}
\caption{The Architecture of the Hybrid MemNet Model}
\label{fig:fig1}
\end{figure}

In this work, we propose a data-driven, end-to-end enhanced encoder-decoder based deep network that summarizes a news article by extracting salient sentences. Figure~\ref{fig:fig1} shows the architecture of the proposed Hybrid MemNet model.
The model consists of document reader (encoder) and a sentence extractor (decoder). Contrary to Cheng and Lapata~\shortcite{cheng2016neural}'s model where they used an attention based decoder, our model uses attention for both encoder and decoder. Our focus is to learn a better document representation that incorporates local as well as global document features along with attention to sentences to capture the notion of saliency of a sentence. Contrary to the orthodox method of computing sentential features, our model uses neural networks and is a purely data-driven approach. Zhang et al.~\shortcite{zhang2014chinese} and  Kim~\shortcite{kim2014convolutional} have shown the successful use of Convolution Neural Networks (CNN) in obtaining latent feature representation. Hence, our network applies CNN with multiple filters to automatically capture latent semantic features. Then a Long Short Term memory (LSTM) network is applied to obtain a comprehensive set of features known as thought vector. This vector captures the overall abstract representation of a document. We obtain the final document representation by concatenating the document embeddings obtained from Convolutional LSTM (Conv-LSTM) and the document embeddings from memory network. The final unified document embedding along with the embeddings of the sentences are used by the decoder to select salient sentences in a document. We experiment with Conv-LSTM encoder as well as Convolutional Bidirectional LSTM (Conv-BLSTM) encoder.
We summarize our primary contributions below:
\begin{enumerate}
\item We propose a novel architecture to learn better unified document representation combining the features from the memory network as well as the features from convolutional LSTM/BLSTM network.
% \item We show that learned unified document representation takes into account n-gram features, sentence level information and also captures the notion of the summary worthiness of sentences.
% \item We investigate the application of memory network in learning the document representation which incorporates attention to sentences.
% \item We investigate the use of convolutional bidirectional long short term memory network as an encoder to learn better thought vector with rich semantics.
\item We investigate the application of memory network (incorporates attention to sentences) and Conv-BLSTM (incorporates n-gram features and sentence level information) for learning better thought vector with rich semantics.
\item We experimentally show that the proposed method outperforms the basic systems and several competitive baselines. Our model achieves significant performance gain on the DUC 2002 generic single-document summarization dataset.
\end{enumerate}

\noindent
We begin by describing our network architecture in Section~\ref{net-arch} followed by experimental details including corpus details in Section~\ref{exp-det}. We analyze our system against various benchmarks in Section~\ref{res-an} and finally conclude our work in Section~\ref{concl}.

\section{Hybrid MemNet Model}
\label{net-arch}
The primary building blocks of our model are:
\begin{itemize}
    \item \textit{Document Encoder} - captures local (n-grams level) information, global (sentence level) information and the notion of summary worthy sentences
  \item \textit{Decoder} - attention based sequence to sequence decoder.  
\end{itemize}
The final unified document encoding and sentences vectors from convolutional sentence encoder are fed to the decoder model. In this section, we discuss details of the encoder and decoder modules.
%\subsection{Type Changes and {\itshape Special} Characters}
\subsection{Document Encoder}
The idea is to learn a unified document representation that not only incorporates n-gram features and sentence level information but also includes the notion of salience and redundancy of sentences. For this purpose, we sum the document representations vectors learned from Convolutional LSTM (Conv-LSTM; for hierarchical encoding) and MemNet~\cite{sukhbaatar2015end} (for capturing salience and redundancy). Since the unified document embedding is learned from the joint interaction of the above mentioned two models, we refer to this network as Hybrid MemNet.

\subsection*{Sentence Encoder}
Convolution neural networks are used to encode sentences as they have been shown to successfully work for multiple sentence-level classification tasks~\cite{kim2014convolutional}.
Conventional convolution neural network uses convolution operation over various word embeddings which is then followed by a max pooling operation. Suppose, d-dimensional word embedding of the $i^{th}$ word in the sentence is $w_i$ and $w_{i:i+n}$ is the concatenation of word embeddings $w_i, ..., w_{i+n}$. Then, convolution operation over a window of $c$ words using a filter of $\theta_t^{c} \in \mathbb{R}^{m\times cd} $ yields new features with $m$ dimensions. Here, $t$ is the filter index. Convolution operation is written as:

\begin{equation}
	\label{eq1}
    f_i^{c} = tanh(\theta_t^{c} \times w_{i:i+c-1} + b)
\end{equation}
Here b is the bias term. We obtain a feature map $F^c$ by applying filter $\theta_t^{c}$ over all possible window of $c$ words in the sentence of length N.

\begin{equation}
	\label{eq2}
    F^{c} = [f_1^{c}, f_2^{c}, ..., f_{N-c+1}^{c}] 
\end{equation}
\noindent
Our intention is to capture the most prominent features in the feature map hence, we use max-over-time pooling operation~\cite{collobert2011natural} to acquire set of features for a filter of fixed window size. Single feature vector ($s$) can be represented as:

\begin{equation}
	\label{eq3}
    s_{\theta_t^{c}} = max\{F^{c}\}
\end{equation}
\noindent
We use multiple convolution nets with different filter sizes \{1, 2, 3, 4, 5, 6, 7\} to compute a list of embeddings which are summed to obtain the  final sentence vector.

\subsection*{Conv-BLSTM Document Encoder}
Since Recurrent Neural Network (RNN) suffers from vanishing gradient problem over long sequences~\cite{siegelmann1992computational}, we use Long Short-Term Memory (LSTM) network. To obtain hierarchical document encoding, sentence vectors obtained from convolutional sentence encoder are fed to the LSTM. This new representation intuitively captures both local as well as global sentential information. We explore LSTM network as well as Bidirectional LSTM network for our experiments. Experiments show that combination of convolution network and Bidirectional LSTM (BLSTM) performs better in our case. BLSTM exploits future context in the sequence as well which is done by processing the data in both directions.

\subsection*{MemNet based Document Encoder}
We leverage a memory network encoder, inspired from the recurrent attention model to solve question answering and language modeling task~\cite{sukhbaatar2015end}. The model uses an attention mechanism and has been shown to capture temporal context. In our case, it learns the document representation which captures the notion of salience and redundancy of sentences.

We first describe the model that implements a single memory hop operation (single layer) then, we extend it to multiple hops in memory. Consider an input set of sentence vectors $s_1, s_2, ...s_i$, obtained from the sentence encoder for a document D. Let $D^\prime$ be the document representation of D obtained from Conv-LSTM model and $D^{\prime\prime}$ is the document embedding from the MemNet model. The entire set of $\{s_i\}$ are transformed into memory vectors $\{m_i\}$ of dimension $d$ in continuous space, using a learned weight matrix $A$ (of size $d\times v$; where $v$ is the embedding size of a sentence). Similarly, an input document embedding $D^\prime$ is transformed via a learned weight matrix $B$ with the same dimension as $A$ to obtain internal state $u$. We then compute the match between $u$ and each memory $m_i$ by taking inner product followed by softmax as follows.

\begin{equation}
	\label{eq1b}
    p_i = softmax(u^Tm_i)
\end{equation}
Where $softmax(z_i) = e^{z_i} / \sum_{j}^{} e^{z_j}$ and $p$ is the probability vector over the inputs. Each $s_i$ also has a corresponding output vector $c_i$ (using another embedding matrix $C$). The output vector from memory $o$ is computed as the sum over the transformed inputs $c_i$, weighted by the probability vector from the input as follows.
\begin{equation}
	\label{eq2b}
	 o = \sum_{i}^{}p_ic_i
\end{equation}
In the case of multiple layer model to handle $K$ (2 in our case) hop operation, the memory layers are stacked and the input to layer $k+1$ is computed as follows.
\begin{equation}
	\label{eq3b}
    u^{k+1} = u^k + o^k
\end{equation}
Let $D^{\prime\prime}$ be the output obtained from the last memory unit $o^K$. Final unified document representation $D_f$ is obtained by summing up the output from the Conv-BLSTM ($D^\prime$) and the output from the MemNet ($D^{\prime\prime}$).
\begin{equation}
	\label{eq4}
    D_f = D^{\prime} + D^{\prime\prime}
\end{equation}
Intuitively, $D_f$ captures the hierarchical information of a document as well as the notion of worthiness of a sentence.

\subsection{Decoder}
The decoder uses an LSTM to label sentences sequentially keeping in mind the individual relevance and mutual redundancy. Taking into account both the encoded document and the previously labeled sentences, labeling of the next sentence is done. If encoder hidden states are denoted by ($h_1, ..., h_m$) and decoder hidden states are denoted by ($\hat{h}_1, ..., \hat{h}_m$) at time step $t$, then for $t^{th}$  sentence the decoder equations are as follows.
\begin{equation}
	\label{deceqn}
    \hat{h_t} = LSTM(p_{t-1}s_{t-1}, \hat{h}_{t-1})
\end{equation}

\vspace{-2pt}
\begin{equation}
	\label{deceqn1}
    p(y_L(t) = 1|D) = \sigma(MLP(\hat{h}_t : h_t)) 
\end{equation}
\noindent
where $p_{t-1}$ is the degree to which the decoder assumes the previous sentence should be a part of summary and is memorized. $p_{t-1}$ is 1 if system is certain. $y_L$ is sentence's label. Concatenation of $\hat{h}_t$ and $h_t$ is given as input to an $MLP$ (Multi-layer Perceptron).

\section{Experimental Results}
\label{exp-det}
In this section of the paper, we present experimental setup for assessing the performance of the proposed system. We present the details of the corpora used for training, evaluation and give implementation details of our approach.

% \begin{table}
%   \caption{Statistics on the DailyMail corpus}
%   \label{dmail}
%   \begin{tabular}{lccc}
%     \toprule
%     &Training&Validation&Test\\
%     \midrule
% Documents & 193986 & 12147& 10350 \\
% Sentences/document & 28 & 27-28 & 27   \\
% Sentences/summary & 3-4 & 4-5 & 3-4\\
% Avg. words/document & 802 & 794& 786\\
%   \bottomrule
% \end{tabular}
% \end{table}

\subsection{Datasets}
For the purpose of training the model, we use the Daily Mail corpus, which was also used for the task of single document summarization by Cheng and Lapata~\shortcite{cheng2016neural}. Overall, this corpus contains 193,986 training documents, 12,417 validation documents and 10,350 test documents. To evaluate our model, we use standard DUC-2002 single document summarization dataset which consists of 567 documents. We also evaluate our system on 500 articles from the DailyMail test set (with human written highlights as the gold standard). The average byte count for each document is 278 and article-highlight pairs are sampled such that the highlights include a minimum of 3 sentences.

\subsection{Implementation Details}
We use top three high-scored sentences subject to the standard word limit of 75 words to generate summaries. The size of the embeddings for word, sentence, and document are set to 150, 300, and 750 respectively. A list of kernel sizes $\{1, 2, 3, 4, 5, 6, 7\}$ is used for convolutional sentence encoder. Two hop operation is performed in the case of MemNet encoder. All LSTM parameters were randomly initialized over a uniform distribution within [-0.05, 0.05]. We use batch size of 20 documents with learning rate 0.001 and the two momentum parameters as 0.99 and 0.999. We use Adam~\cite{kingma2014adam} as optimizer.
%Pre-trained 150-dimensional word2vec~\cite{mikolov2013distributed} embeddings are used to initialize word vectors.

\subsection{Evaluation Metrics}
We evaluate the quality of system summaries using ROUGE~\cite{lin2003automatic}: ROUGE-1 (unigram overlap), ROUGE-2 (bigram overlap) as means of assessing informativeness and ROUGE-L as means of assessing fluency.
%We present the evaluation report on both DUC-2002 and DailyMail dataset in Table~\ref{res}.

\subsection{Baseline Methods}
\label{baseline}
We evaluate our system against several state-of-the-art baselines. We select best systems having state-of-the-art summarization results on DUC 2002 corpus for single document summarization task, which are: (1) ILP~\cite{woodsend2010automatic}, (2) TGRAPH~\cite{parveen2015topical}, (3) URANK~\cite{wan2010towards}, (4) NN-SE~\cite{cheng2016neural}, (5) SummaRuNNer~\cite{nallapati2016summarunner}, and (6) Deep-Classifier~\cite{nallapati2016classify}. ILP is a phrase-based extraction model that selects salient phrases and recombines them subject to length and grammar constraints via Integer Linear Programming (ILP). TGRAPH is a graph-based sentence extraction model. URANK uses a unified ranking for single- as well as multi-document summarization. We also use LEAD as a standard baseline of simply selecting the leading three sentences from the document as the summary. NN-SE is a neural network \vspace{0pt} based sentence extractor. Deep-Classifier uses GRU-RNN to sequentially accept or reject each sentence in the document for being in summary. SummaRuNNer is an RNN based extractive summarizer.

\section{Results and Analysis}
\label{res-an}
In this section, we compare the performance of our system against summarization baselines mentioned in Section~\ref{baseline}. Table~\ref{res} shows our results on the DUC 2002 test dataset and on the 500 samples from the Daily Mail corpus. Hybrid MemNet
%\footnote{We make our implementation open-sourced at \url{https://anonymous.com/examplecode}}
represents our system with Conv-LSTM encoder and MemNet encoder, while Hybrid MemNet$^{*}$ uses Conv-BLSTM encoder and MemNet encoder. It is evident from the results that our system (Hybrid MemNet/ Hybrid MemNet$^*$) outperforms the LEAD and ILP baselines with a large margin which is an encouraging result as our system does not have access to manually-crafted features, syntactic information and sophisticated linguistic constraints as in the case of ILP. Results also show that our system performs better without the sentence ranking mechanism (URANK). It also achieves significant performance gain against NN-SE, Deep-Classifier, and SummaRuNNer.

To explore the contribution of the MemNet encoder towards the performance of our system we compare results of NN-SE with Hybrid MemNet. Note that there is significant performance gain of about 2\% in the results. Post-hoc Tukey tests showed that the proposed Hybrid MemNet model is significantly ($p < 0.01$) better than NN-SE. This is due to the fact that MemNet learns document representation which captures salience estimation of a sentence (using the attention mechanism) prior to the summary generation. We also notice that replacing LSTM with BLSTM in the encoder improves the performance of the system. This may be because BLSTM in our setting is able to learn a richer set of semantics as they exploit some notion of future context as well by processing the sequential data in both directions, while LSTM is only able to make use of the previous context.

\begin{table}
\small
  \caption{Rouge Evaluation (\%) on the DUC-2002 Corpus and 500 Samples from the Daily Mail Corpus}
  \label{res}
  \begin{tabular}{llll}
    \toprule
    DUC 2002&ROUGE-1&ROUGE-2&ROUGE-L\\
    \midrule
    LEAD & 43.6 & 21.0 & 40.2\\
    ILP & 45.4& 21.3& 42.8\\
    TGRAPH & 48.1 & 24.3 & --\\
    URANK & 48.5 & 21.5 & --\\
    NN-SE & 47.4 & 23.0 & 43.5\\
    Deep-Classifier & $46.8\pm0.9$& $22.6\pm0.9$& $43.1\pm0.9$\\
    SummaRuNNer & $46.6\pm0.8$&$23.1\pm0.9$ & $43.03\pm0.8$\\
    Hybrid MemNet & 49.1 & 24.7 & 44.6\\
    Hybrid MemNet$^{*}$ & \textbf{50.1} & \textbf{25.2} & \textbf{44.9}\\
  \bottomrule
\end{tabular}

	\begin{tabular}{llll}
    \toprule
    DailyMail&ROUGE-1&ROUGE-2&ROUGE-L\\
    \midrule
    LEAD & 20.4 & 7.7 & 11.4\\
    NN-SE & 21.2 & 8.3 & 12.0\\
    Deep-Classifier & $26.2\pm0.4$& $10.7\pm0.4$& $14.4\pm0.4$\\
    SummaRuNNer & $26.2\pm0.4$&$10.8\pm0.3$ & $14.4\pm0.3$\\
    
    Hybrid MemNet & 27.1 & 11.6 & 15.2\\
    Hybrid MemNet$^{*}$ & \textbf{27.9} & \textbf{12.2} & \textbf{15.5}\\
  \bottomrule
\end{tabular}
\end{table}
\vspace{-1pt}

\section{Conclusions}
\label{concl}
In this work, we proposed a data-driven end-to-end deep neural network approach for extractive summarization of a document. Our system makes use of a combination of memory network and convolutional bidirectional long short term memory network to learn better unified document representation which jointly captures n-gram features, sentence level information and the notion of the summary worthiness of sentences eventually leading to better summary generation. Experimental results on DUC 2002 and Daily Mail datasets confirm that our system outperforms several state-of-the-art baselines.

\bibliographystyle{ACM-Reference-Format}
\bibliography{sigproc} 

\end{document}